\documentclass[11pt]{article}

\usepackage[final]{acl}

\usepackage{times}
\usepackage{latexsym}
\usepackage[T1]{fontenc}

\usepackage[utf8]{inputenc}

\usepackage{microtype}

\usepackage{inconsolata}

\usepackage{graphicx}
\usepackage{array} 
\usepackage{multirow}
\usepackage{booktabs}
\usepackage{amsmath} 
\usepackage{amssymb}
\newcommand{\pmstd}[2]{#1{ \scriptsize  $\pm$ #2}}
\usepackage{arydshln} 
\usepackage[noabbrev,capitalise]{cleveref}
\usepackage{xcolor} 
\usepackage{url}
\usepackage{float}
\usepackage{placeins}
\usepackage{fontawesome5}
\usepackage{hyperref}
\usepackage{algorithm}
\usepackage{algpseudocode}

\usepackage{tcolorbox}
\usepackage{xcolor}

\usepackage{algorithm}
\usepackage{algpseudocode}
\usepackage{listings}
\usepackage{booktabs}
\usepackage[table,xcdraw]{xcolor}
\usepackage{amsmath}
\definecolor{codegray}{rgb}{0.98,0.98,0.98}
\tcbset{
  myalgo/.style={
    colback=white, 
    colframe=white, 
    sharp corners,
    boxrule=0.5pt,
    left=0mm,
    right=2mm,
    top=0mm,
    bottom=1mm,
  }
}
\definecolor{keywordcolor}{rgb}{0.0,0.0,0.5}     
\definecolor{stringcolor}{rgb}{0.58,0.0,0.0}      
\definecolor{commentcolor}{rgb}{0.25,0.5,0.25}    

\definecolor{hatecolor}{RGB}{205,92,92}    %
\definecolor{toxiccolor}{RGB}{255,120,40}   %
\definecolor{neutralcolor}{RGB}{0,128,255} %

\title{Modeling Human Perspectives with Socio-Demographic Representations}

\author{Leixin Zhang \\
  University of Tübingen  \\
  \texttt{leixin.zhang@uni-tuebingen.de} \\\And
  Çağrı Çöltekin \\
  University of Tübingen \\
  \texttt{cagri.coeltekin@uni-tuebingen.de} \\ 
  }

\begin{document}
\maketitle
\begin{abstract}

Humans often hold different perspectives on the same issues. In many NLP tasks, annotation disagreement can reflect valid subjective perspectives. Modeling annotator perspectives and understanding their relationship with other human factors, such as socio-demographic attributes, have received increasing attention. Prior work typically focuses on single demographic factors or limited combinations. However, in real-world settings, annotator perspectives are shaped by complex social contexts, and finer-grained socio-demographic attributes can better explain human perspectives. In this work, we propose Socio-Contrastive Learning, a method that jointly models annotator perspectives while learning socio-demographic representations. Our method provides an effective approach for the fusion of socio-demographic features and textual representations to predict annotator perspectives, outperforming standard concatenation-based methods. The learned representations further enable analysis and visualization of how demographic factors relate to variation in annotator perspectives.
Our code is available at \faGithub\, \href{https://github.com/Leixin-Zhang/Socio_Contrastive_Learning}{GitHub}.\footnote{\href{https://github.com/Leixin-Zhang/Socio_Contrastive_Learning}{\url{https://github.com/Leixin-Zhang/Socio_Contrastive_Learning}}}

\end{abstract}

\section{Introduction}\label{intro}

Recent studies have shown that human disagreement is widespread across many annotation tasks. Some instances lack a single ground truth and annotation variation reflects diverse but valid perspectives \cite{plank-2022-problem,cabitza2023toward,leonardelli2023semeval,wang2023collective,pham-etal-2023-solving}.

This phenomenon is particularly evident in subjective tasks, such as hate speech and offensive content detection \cite{sang2022origin,waseem-2016-racist}. Some studies also suggest that individuals’ perspectives are associated with their lived experiences and cultural contexts. For example, \citet{sap-etal-2022-annotators} find that conservative annotators are more likely to rate African American English as toxic. \citet{larimore-etal-2021-reconsidering} show that white and non-white annotators differ significantly in their ratings of racist language. 

Despite these advances, pitfalls remain in analyzing the role of socio-demographic features in perspective modeling. Prior research often examines annotator subgroups using a single demographic attribute at a time, such as gender, and analyzes the statistical difference between its subcategories, such as female and male. However, this approach is limited, as a single attribute cannot fully capture the complex interplay between social contexts and the formation of human perspectives. Instead, richer combinations of socio-demographic attributes are more likely to reflect the nuanced experiences and social context, and to better capture the diversity of annotators’ perspectives.
In this work, we adopt a machine learning approach to analyze annotators’ socio-demographic attributes and address the following research questions:
\vspace{-5pt}

\begin{itemize}
    \item \textsc{Research Question 1:} To what extent do socio-demographic features contribute to perspective prediction in hate speech and toxic content tasks, and which features are most predictive of annotators’ perspectives? \vspace{-4pt}
    \item \textsc{Research Question 2: } How should socio-demographic features be encoded and fused with textual representations to improve model performance?
\end{itemize}
\vspace{-5pt}

 As the key contribution of this work, we propose \textbf{Socio-Contrastive Learning}, a novel architecture that jointly learns (i) annotators' socio-demographic representations inferred from their labeling behavior and (ii) socio-demographic-specific label predictions. Our method outperforms existing models on label prediction while simultaneously producing annotators' socio-demographic representations that enable analysis of their association with perspective variation.

\section{Related Studies}

This section presents prior work on modeling individual perspectives. \cref{individual_modeling} focuses on methods for predicting individual annotators' labels without using socio-demographic features, while \cref{socio-enriched_learning} focuses on approaches that incorporate socio-demographic information into the learning process.

\subsection{Individual Perspective Modeling}\label{individual_modeling}

Modeling each annotator's labels directly, rather than aggregating via majority vote, has been shown to improve performance \cite{davani-etal-2022-dealing,uma2021learning,mokhberian-etal-2024-capturing,zhang2026disagreement}.

\citet{kanclerz2021controversy} propose personalized approaches that leverage annotator label embeddings derived from their partial annotations, showing that even a small number of annotations on highly controversial data can significantly outperform generalized models. \citet{davani-etal-2022-dealing} introduce three methods for modeling annotation variation: an \textit{ensemble} approach that aggregates predictions from annotator-specific models, a \textit{multi-label} approach that represents annotators’ labels as a target vector, and a \textit{multi-task} approach that assigns a separate prediction head to each annotator. \citet{mokhberian-etal-2024-capturing} use an embedding layer to represent individual annotators for annotator-specific prediction.

\subsection{Socio-Demographics Enriched Modeling}\label{socio-enriched_learning}
Some prior studies show that certain socio-demographic features of annotators are associated with their annotated labels.
\citet{huang2023culturally} identify differences in culture-related natural language inference judgments between annotators from the United States and India. 
In the context of socio-demographic enriched learning, \textit{Jury Learning} \cite{gordon2022jury} models annotator-specific labels by concatenating socio-demographic features and annotator IDs with text representations. Final predictions are aggregated via a sampling process with a predefined demographic composition, allowing practitioners to control which groups’ perspectives are reflected in the final outputs and in what proportions.

\citet{orlikowski-etal-2023-ecological} investigate the influence of socio-demographic features by grouping annotators according to a single attribute and employing group-specific layers for each subcategory (e.g., Age: 25–34, 35-44, etc.).
The results do not show performance improvements over a baseline model without socio-demographic features and randomly shuffled attribute groups, and they conclude that annotation patterns cannot be explained by socio-demographic attributes.
However, using the same dataset for toxic content tasks, our study arrives at a conclusion different from that in \citet{orlikowski-etal-2023-ecological}. We find that incorporating richer combinations (instead of an individual attribute in isolation) reveals a substantial contribution of socio-demographic information to modeling and explaining annotation variation.

\section{Methodology}
To address the first research question: the extent to which socio-demographic features contribute to perspective prediction, we compare model architectures in the following two categories:

\paragraph{Category 1:} Models that do not use socio-demographic features:

\begin{itemize}

    \item \textbf{Simple Model:} This model uses only text embeddings to predict aggregated labels. When a text item is annotated by multiple annotators, their labels are aggregated into a single ground-truth label via majority voting.

    \item \textbf{Multi-Task Model:} This model uses text embeddings as input but predicts each annotator’s label in a multi-task setup as \citet{davani-etal-2022-dealing}. The architecture consists of shared layers for all annotators, followed by separate output heads for each annotator.

\end{itemize}

\paragraph{Category 2:} Models that leverage socio-demographic features.
We experiment with two commonly used methods and introduce a novel approach for encoding annotators' socio-demographic attributes.

\begin{figure}[!t]
    \centering
    \includegraphics[width=1\linewidth]{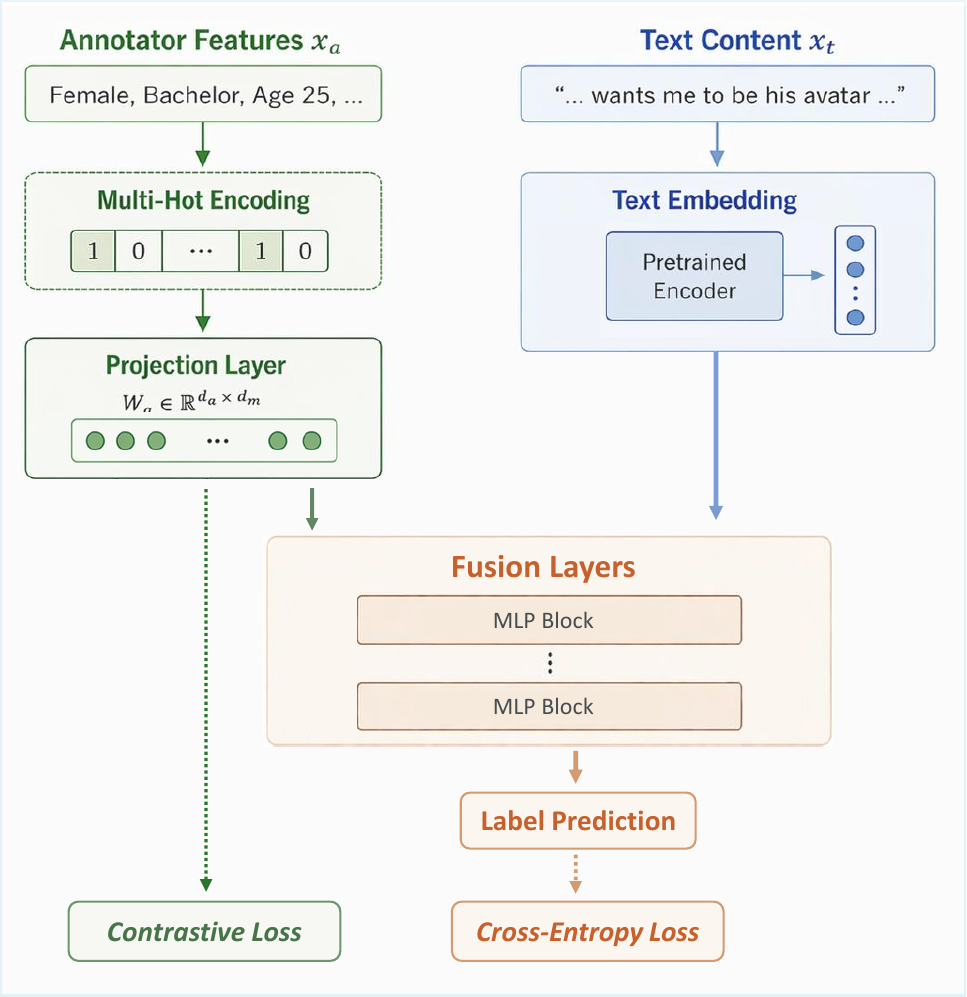}
        \caption{Socio-Contrastive Model Architecture}
        \label{fig:model_arch}
\end{figure}

\begin{itemize}
    \item \textbf{Multi-Hot Encoding:} Annotators’ socio-demographic features are encoded as multi-hot vectors and concatenated with text embeddings as model input.

    \item \textbf{Socio-Demographic Embedding:} 
    Socio-demographic features are encoded using the same pretrained model as the text encoder, producing representations in a shared vector space for concatenation.

    \item \textbf{Socio-Contrastive Learning (Our Method):} 
    As shown in \cref{fig:model_arch}, multi-hot socio-demographic encodings are first projected into a learnable space (Projection Layer) and optimized with a contrastive loss to capture annotator-specific patterns. The resulting representations are then concatenated with text embeddings for label prediction.

\end{itemize}

\section{Socio-Contrastive Learning} \label{our_model}

The primary objective of the model is to predict annotator labels. Meanwhile, the model refines socio-demographic representations using a contrastive loss based on each annotator’s labels, enabling them to capture socio-demographic–relevant patterns.

\subsection{Motivation}

We hypothesize that, for certain subjective tasks, an individual’s perspective—reflected in their annotations on text items—is associated with the socio-demographic groups to which they belong. In other words, prediction accuracy of annotator labels is expected to improve when fine-grained socio-demographic attributes are available. Formally, let $\hat{y} = \arg\max_y P(y \mid \cdot)$ denote the predicted label and $y$ the ground-truth label. We posit that:
\[
\mathbb{P}(\hat{y} = y \mid \text{text}, \text{socio-demo}) > \mathbb{P}(\hat{y} = y \mid \text{text})
\]

\noindent We expect that socio-demographic attributes are not equally predictive of annotators' perspectives. Certain attributes may be more closely associated with specific judgment patterns, while others may contribute weaker signals. To amplify the most informative attributes, we apply contrastive learning to refine socio-demographic representations based on annotation similarity, encouraging representations to be closer for annotators with similar annotation patterns and farther apart otherwise.

\subsection{Model Design}

The model input consists of two components: a multi-hot encoding of the annotator’s socio-demographic attributes and a text representation encoded by a pretrained model. Before concatenation, the multi-hot socio-demographic vector is  
projected into a learnable space and optimized with a contrastive loss to capture annotator-specific patterns. 
The primary task of the model is to predict each annotator's label, using a cross-entropy loss. In parallel, the model also learns to optimize the socio-demographic representations through a contrastive loss, which guides the representations to capture annotation patterns.

\subsection{Contrastive Representation Learning}\label{contrastive_loss}

We apply contrastive loss to learn annotators' socio-demographic representations. For a given text, annotators who provide the same label are treated as \textbf{positive cases}, and annotators who provide different labels serve as \textbf{negative cases}. 

Standard contrastive learning methods, such as the InfoNCE loss, typically rely on predefined positive and negative cases and assume a fixed number of such examples (e.g., pairs or triplets). However, crowdsourced datasets pose several challenges. Texts are annotated by a variable number of annotators, ranging from a few to over 100, resulting in inconsistent numbers of positive and negative pairs across instances. In addition, items with perfect agreement contain no negative pairs. Excluding such instances would discard useful data and reduce the effectiveness of perspective modeling.

To address these issues, we design a flexible contrastive learning scheme tailored for socio-demographic representation learning. Our approach performs contrastive learning within each batch, prioritizing samples that share the same text ID (the same text annotated by multiple annotators) into the same batch. If the number of annotations for a given text is smaller than the batch size, other texts are included to fill the batch; if it exceeds the batch size, the remaining annotations are processed in subsequent batches. During training, only annotations of the same text are considered in the contrastive loss, while those from different texts are masked out.

Within a batch, for negative pairs (annotators who disagree on a label), the loss is weighted by the similarity of their socio-demographic representations, pushing them apart. For positive pairs (annotators who agree on a label), the loss is weighted by the dissimilarity of their representations, pulling them closer. This scheme allows the model to learn socio-demographic representations that reflect systematic differences and agreements in annotation patterns. A detailed description of the algorithm is provided in \cref{contrastive_loss_algorithm}.

\begin{table}[t]
\renewcommand{\arraystretch}{1.2}

\resizebox{\columnwidth}{!}{
\begin{tabular}{c lccc r}
\toprule
\textbf{} & \textbf{Data Split} & \textbf{Text} & \textbf{Uniq. Text} & \textbf{Ann.} & \textbf{Labels} \\
\midrule
\multirow{4}{*}{\rotatebox{90}{\textsc{HateSpeech}}} 
    & \multirow{2}{*}{Train (70\%)} & \multirow{2}{*}{22,942} & \multirow{2}{*}{6,227} & \multirow{2}{*}{2,315} & 10,179 {H} \\
    &                              &                          &                          &                          & 12,763 {N} \\
    \cmidrule{2-6}
    & \multirow{2}{*}{Test (30\%)}  & \multirow{2}{*}{10,359} & \multirow{2}{*}{2,670} & \multirow{2}{*}{2,263} & 4,415 {H} \\
    &                              &                          &                          &                          & 5,944 {N} \\
\midrule
\multirow{4}{*}{\rotatebox{90}{\textsc{Toxic}}} 
    & \multirow{2}{*}{Train (60\%)} & \multirow{2}{*}{25,556} & \multirow{2}{*}{4,638} & \multirow{2}{*}{3,517} & 10,895 {T} \\
    &                              &                          &                          &                          & 14,661 {N} \\
    \cmidrule{2-6}
    & \multirow{2}{*}{Test (40\%)}  & \multirow{2}{*}{16,151} & \multirow{2}{*}{3,092} & \multirow{2}{*}{2,656} & 7,210 {T} \\
    &                              &                          &                          &                          & 8,914 {N} \\
\bottomrule
\end{tabular}}
\vspace{10pt}

\footnotesize{\textbf{H}: {Hate Speech}; \textbf{T}: {Toxic}; \textbf{N}: {Not Hate Speech or Toxic}. \vspace{5pt}

\textbf{Ann.}:  Annotator Size. \vspace{5pt}\\Train set and test set splits are performed using unique text IDs to prevent leakage between training and evaluation. }
\caption{Dataset statistics for the hate speech and toxic content classification tasks.}%
\label{tab:dataset_stats}
\end{table}
\begin{table*}[h]
\centering
\renewcommand{\arraystretch}{1.1}
\resizebox{\textwidth}{!}{%
\begin{tabular}{lcccccc}
\toprule
\multirow{2}{*}{\textbf{Model}}& \multicolumn{3}{c}{\textbf{Hate Speech}} 
& \multicolumn{3}{c}{\textbf{Toxic}} \\ \cmidrule(lr){2-4} \cmidrule(lr){5-7}

& {Precision} & {Recall} & {F1}
& {Precision} & {Recall} & {F1} \\ \hline
Simple Model
& 0.438 {\scriptsize$\pm$ 0.046}
& 0.395 {\scriptsize$\pm$ 0.065}
& 0.415 {\scriptsize$\pm$ 0.052}
& 0.453 {\scriptsize$\pm$ 0.005}
& 0.525 {\scriptsize$\pm$ 0.041}
& 0.486 {\scriptsize$\pm$ 0.019} \\

Multi-Task
& 0.670 {\scriptsize$\pm$ 0.028}
& 0.565 {\scriptsize$\pm$ 0.108}
& 0.608 {\scriptsize$\pm$ 0.074}
& 0.614 {\scriptsize$\pm$ 0.005}
& 0.487 {\scriptsize$\pm$ 0.012}
& 0.543 {\scriptsize$\pm$ 0.006} \\ \hdashline

Socio Multi-Hot
& 0.759 {\scriptsize$\pm$ 0.009}
& 0.629 {\scriptsize$\pm$ 0.042}
& 0.687 {\scriptsize$\pm$ 0.026}
& 0.626 {\scriptsize$\pm$ 0.011}
& 0.607 {\scriptsize$\pm$ 0.023}
& 0.616 {\scriptsize$\pm$ 0.009} \\

Socio Embedding
& 0.750 {\scriptsize$\pm$ 0.013}
& 0.655 {\scriptsize$\pm$ 0.031}
& 0.699 {\scriptsize$\pm$ 0.015}
& 0.666 {\scriptsize$\pm$ 0.015}
& 0.596 {\scriptsize$\pm$ 0.036}
& 0.628 {\scriptsize$\pm$ 0.013} \\

Socio Contrastive (Ours)
& 0.729 {\scriptsize$\pm$ 0.037}
& 0.727 {\scriptsize$\pm$ 0.068}
& \textbf{0.725} {\scriptsize$\pm$ 0.018}
& 0.625 {\scriptsize$\pm$ 0.013}
& 0.667 {\scriptsize$\pm$ 0.027}
& \textbf{0.645} {\scriptsize$\pm$ 0.008} \\

Ablation (w/o $\mathcal{L}_{contrastive}$)
& 0.743 {\scriptsize$\pm$ 0.018}
& 0.631 {\scriptsize$\pm$ 0.051}
& {0.681} {\scriptsize$\pm$ 0.030}
& 0.649 {\scriptsize$\pm$ 0.009}
& 0.621 {\scriptsize$\pm$ 0.024}
& {0.634} {\scriptsize$\pm$ 0.009} \\
\bottomrule
\end{tabular}}
\caption{Performance comparison of five models on hate speech and toxicity classification (mean ± standard deviation). The best F\textsubscript{1} scores are highlighted in \textbf{bold}. Models above the dashed line do not use socio-demographic features, and models below the dashed line incorporate socio-demographic features. Ablation results (w/o $\mathcal{L}_{contrastive}$) indicate the removal of the contrastive loss from our method.}
\label{overall_result_table}

\end{table*}

\section{Experiments}

This section presents the datasets and implementation details used in our experiments.
\subsection{Datasets}

We conduct our study using two crowd-annotated datasets that include rich socio-demographic metadata about the annotators: a hate speech dataset \cite{kennedy2020constructing} and a toxicity dataset \cite{kumar2021designing}. Prior work \cite{gordon2022jury,orlikowski-etal-2023-ecological} has documented substantial annotation disagreement in both tasks.

Both datasets provide the following socio-demographic attributes of annotators: education, political ideology, age, gender, race, and sexuality. Additionally, the toxicity dataset includes the annotator's income range and self-reported importance of religions. The hate speech dataset includes whether the annotator is a parent.
These variables serve as the socio-demographic features used in our modeling.

For reliability, we remove items with too few annotators (fewer than 2 for hate speech, and fewer than 4 for the toxic dataset) as well as annotations from annotators who contributed an insufficient number of labels (fewer than 20 for both datasets). After preprocessing, the hate speech dataset contains 6,227 unique texts and the toxicity dataset contains 4,638 unique texts. On average, each text is annotated by \textasciitilde
{4} annotators for hate speech and \textasciitilde
6 annotators for toxic detection. Both datasets were originally annotated using Likert scales ratings. For direct comparison with other baseline models, we convert them to binary labels by mapping a score of 0 to the non-hate or non-toxic class and any score above 0 to the hate or toxic class. \cref{tab:dataset_stats} presents detailed statistics for both datasets.

\subsection{Implement Details}

All models are implemented with {PyTorch} \cite{paszke2019pytorch}.

\paragraph{Text Representations.}

Text inputs are encoded under three settings with three pretrained encoders:
(i) Sentence-BERT \cite{reimers-2019-sentence-bert}; (ii) \textsc{BERT} \cite{devlin-etal-2019-bert} and (iii) RoBERTa \cite{liu2019RoBERTa}. 

\paragraph{Model Architecture and Training.}
Two dense layers with dropout are applied across all models. Cross-entropy serves as the primary training objective. For the Socio-Contrastive model, a customized contrastive loss is combined with cross-entropy, with both objectives weighted equally. We further conduct an ablation study by setting the contrastive loss weight to 0 for comparison. Model parameters are optimized using the Adam optimizer \cite{kingma2014adam}.

\paragraph{Evaluation Protocol.}

Each model is trained under multiple hyperparameter configurations to determine the optimal setup (details are provided in \cref{parameter}). After selecting the optimal hyperparameters, we train each model for six independent runs and report the mean performance and standard deviation across these runs. 
Given the presence of annotation disagreement and our focus on modeling divergent annotator perspectives, we evaluate models based on their ability to predict individual annotator labels rather than aggregated or majority-vote labels.
A threshold of 0.5 on the sigmoid outputs is used to assign binary labels (hate vs.\ non-hate, toxic vs.\ non-toxic) and compute precision, recall, and F\textsubscript{1}. We additionally present the \textsc{auc-roc}, which assesses the model’s ranking ability by measuring how well it separates positive and negative instances across all possible decision thresholds.

\section{Results and Discussion}

\begin{figure}[t]
    \centering
    \includegraphics[width=\linewidth]{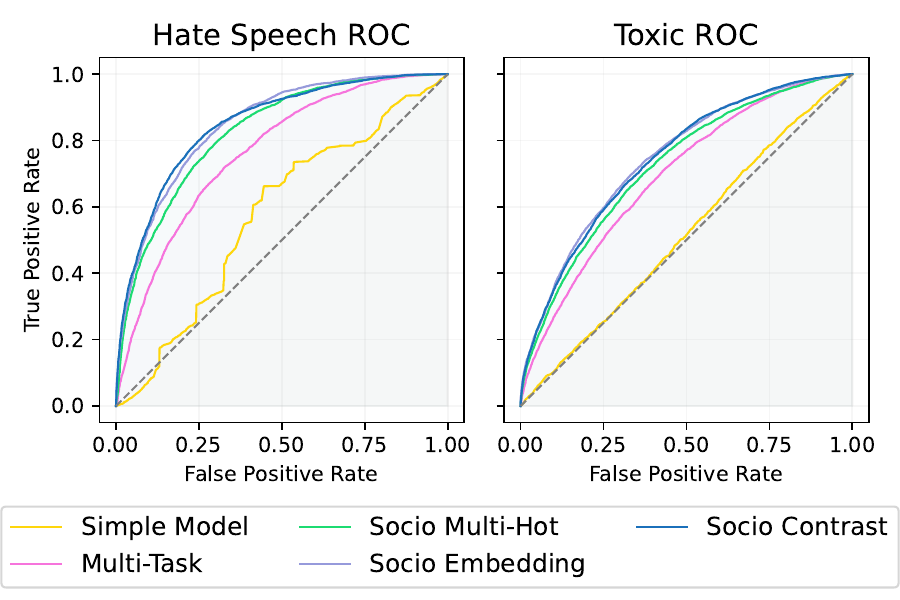}
        \caption{\textsc{roc} curve: the prediction performance of five models}
        \label{fig:roc_curve}
\end{figure}

Models that leverage socio-demographic information consistently outperform those that rely on textual input alone, as evidenced by the F\textsubscript{1} scores in \cref{overall_result_table} and \textsc{roc} curves in \cref{fig:roc_curve}.\footnote{We report results using the best-performing pre-trained encoder in the main text. Sentence-BERT outperforms BERT and RoBERTa for text encoding, which aligns with the results from other studies \cite{reimers-2019-sentence-bert,zhang-coltekin-2024-tubingen,zhang-etal-2024-unveiling}. Results with BERT and RoBERTa encoders are provided in \cref{bert_robert_results}.}

\subsection{Text-Only Models}

Simple model uses aggregated labels as training targets, perform poorly in predicting individual annotator labels. This result highlights the importance of modeling unaggregated annotations to capture diverse perspectives. 
The multitask architecture proposed by \citet{davani-etal-2022-dealing}, which predicts each annotator’s labels via a separate output head, underperforms socio-demographic–enriched models. This stems from the need for sufficient per-annotator data to effectively train annotator-specific heads. However, crowdsourced datasets typically involve a large number of annotators, each contributing only a small number of annotations (often around 20 or fewer after data splitting). Under such conditions, the multitask architecture is not well suited to settings with many annotators but limited per-annotator data. This limitation also explains the findings of \citet{orlikowski-etal-2023-ecological} who report no performance gains from incorporating socio-demographic features. 
Their approach adopts a similar per-annotator head design, which, under limited per-annotator data conditions, constrains the model from learning meaningful annotation patterns. In addition, assigning an independent output head to each annotator introduces substantial computational overhead, particularly given that both datasets contain over 2,000 annotators. 


\begin{figure*}[t]
    \centering
    \includegraphics[width=1\linewidth]{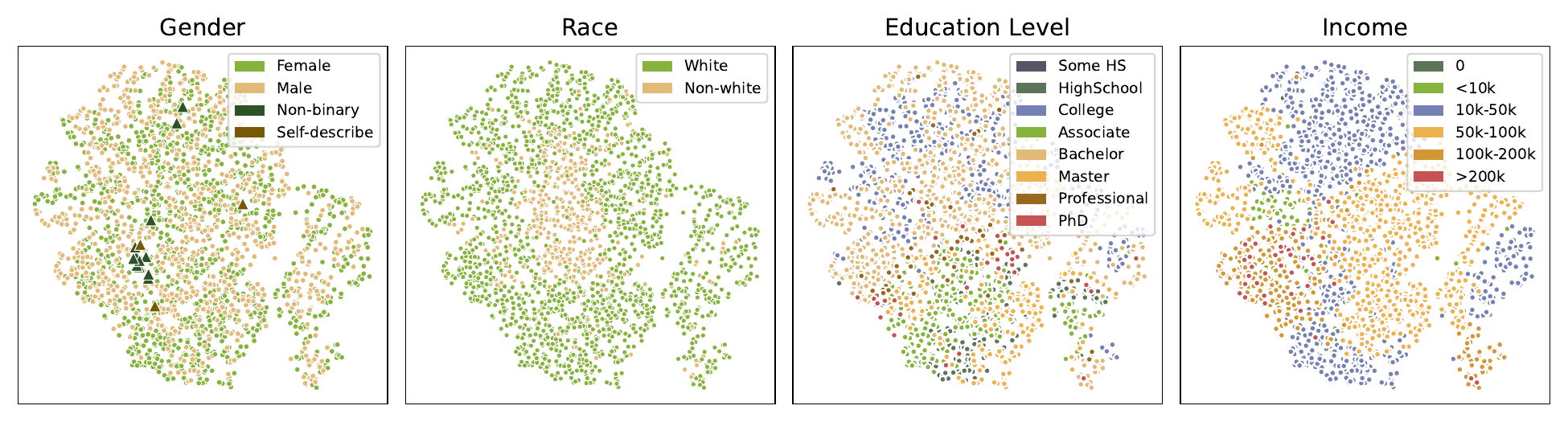}
    \caption{Visualization of Contrastively Learned Socio-Demographic Representations for \textbf{Hate Speech} Task}
        \label{fig:hatesppeech_rep}
\end{figure*}

\begin{figure*}[t]
    \centering
    \includegraphics[width=1\linewidth]{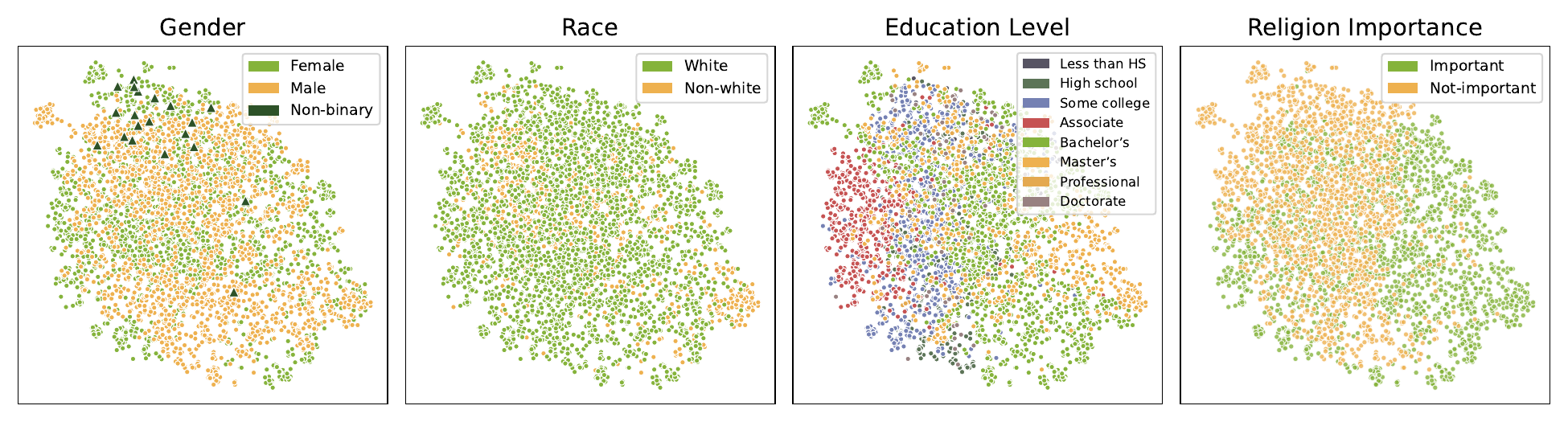}
    \caption{Visualization of Contrastively Learned Socio-Demographic Representations for the \textbf{Toxic} Task}
    \label{fig:toxic_rep}
\end{figure*}

\vspace{10pt}

\subsection{Socio-Demographic Enriched Models}

Our results suggest that richer combinations of socio-demographic attributes provide more informative signals for predicting hate speech and toxic content opinions, particularly under the Socio-Contrastive Learning setting.

The contrastively trained socio-demographic model achieves the best overall performance, reaching an F\textsubscript{1} score of 0.725 on the hate speech dataset and 0.645 on the toxicity dataset. Ablation results further demonstrate the effectiveness of contrastive learning, with performance drops by 4.4\% and 1.1\%, respectively, when the contrastive loss is removed. This can be explained by the fact that the contrastive strategy strengthens socio-demographic signals that are more closely associated with annotation patterns and improves the interaction between socio-demographic information and textual representations. It produces dense socio-demographic representations that better capture annotation patterns and are effectively adapted to the task through fusion with text representations.

The socio-embedding model, which encodes socio-demographic attributes using pretrained Sentence-BERT (\cref{overall_result_table}) or RoBERTa (\cref{appendix_result_table_RoBERTa} in \cref{bert_robert_results}), outperforms the corresponding multi-hot encoding model using the same encoder. This suggests that, for these two pre-trained models, mapping socio-demographic attributes into the aligned text embedding space better captures their interaction with textual content.
However, for BERT-encoded texts, multi-hot encoding outperforms BERT-based socio-demographic representations (\cref{appendix_result_table_bert} in \cref{bert_robert_results}).

\section{Socio-Demographic Contributions}

Our Socio-Contrastive model offers an additional advantage by learning the representations of the full combination of an annotator’s socio-demographic features. As these vectors are optimized based on annotators’ labeling behavior, annotators who consistently show disagreement are pushed farther apart in the representation space during training, while those with similar annotation patterns are brought closer together. The resulting distances between annotator vectors serve as an interpretable signal of \textit{perspective divergence}, enabling analysis of how specific socio-demographic attributes contribute to differences in annotation behavior. To examine these contributions, we employ two approaches:
(1) visualization of the learned representations in \cref{sec:visualization}, and
(2) statistical analysis of distances across socio-demographic groups in \cref{sec:statistical_analysis_of_vectors}.

\begin{table*}[t]
\renewcommand{\arraystretch}{1.14}
\resizebox{\textwidth}{!}{
\begin{tabular}{lcccccc}
\toprule
\multirow{2}{*}{\textbf{Attribute}} & 
\multicolumn{3}{c}{\textbf{Hate Speech}} & 
\multicolumn{3}{c}{\textbf{Toxic Content}} \\
\cmidrule(lr){2-4} \cmidrule(lr){5-7}
& {Observed} & {Random} & {Ratio} & {Observed} & {Random} & {Ratio} \\
\midrule
Education & \pmstd{0.631}{0.007} & \pmstd{0.244}{0.005} & \pmstd{2.590}{0.045} & \pmstd{0.702}{0.005} & \pmstd{0.246}{0.004} & \pmstd{2.857}{0.045} \\
Political Ideology & \pmstd{0.415}{0.006} & \pmstd{0.162}{0.002} & \pmstd{2.566}{0.037} & \pmstd{0.794}{0.004} & \pmstd{0.347}{0.002} & \pmstd{2.285}{0.019} \\
Income & \pmstd{0.710}{0.006} & \pmstd{0.343}{0.005} & \pmstd{2.070}{0.024} & - & - & - \\
Age Group & \pmstd{0.466}{0.005} & \pmstd{0.232}{0.004} & \pmstd{2.015}{0.030} & \pmstd{0.678}{0.005} & \pmstd{0.251}{0.004} & \pmstd{2.701}{0.035} \\
Religion Importance & - & - & - & \pmstd{0.882}{0.003} & \pmstd{0.506}{0.002} & \pmstd{1.743}{0.008} \\
Gender & \pmstd{0.827}{0.004} & \pmstd{0.503}{0.004} & \pmstd{1.645}{0.013} & \pmstd{0.823}{0.004} & \pmstd{0.496}{0.001} & \pmstd{1.659}{0.008} \\
Parental Status & - & - & - & \pmstd{0.820}{0.003} & \pmstd{0.501}{0.001} & \pmstd{1.636}{0.007} \\ 
Race & \pmstd{0.827}{0.005} & \pmstd{0.685}{0.010} & \pmstd{1.208}{0.012} & \pmstd{0.829}{0.005} & \pmstd{0.589}{0.009} & \pmstd{1.409}{0.018} \\
Sexuality & \pmstd{0.903}{0.003} & \pmstd{0.777}{0.010} & \pmstd{1.162}{0.013} & \pmstd{0.926}{0.003} & \pmstd{0.762}{0.009} & \pmstd{1.216}{0.012} \\ \hdashline
Transgender & \pmstd{0.968}{0.002} & \pmstd{0.974}{0.005} & \pmstd{0.994}{0.003} & - & - & - \\
\bottomrule
\end{tabular}}
\vspace{3pt}\\ \small 
\textit{Note:} Values are presented as mean ± standard deviation.  
``Observed'' = the observed probability that nearest neighbors share the same attribute.  
``Random'' = the expected probability by chance.  
``Ratio'' = Observed / Random.  
Above the dashed line: Ratio > 1; Below the dashed line: Ratio < 1.  
\caption{Analysis of Annotator Socio-Demographics}
\label{tab:homophily_analysis}
\end{table*}

\subsection{Representation Visualization}\label{sec:visualization}
After the contrastive model is trained, we obtain learned vector representations for each unique annotator in the dataset (2,316 in the hate speech data and 4,408 in the toxic data). \\

\noindent UMAP \cite{mcinnes2018umap}\footnote{\url{https://umap-learn.readthedocs.io/en/latest/}}
 is used for dimensionality reduction, projecting annotator vectors into a two-dimensional space.

\paragraph{Hatespeech Dataset}
As shown in \cref{fig:hatesppeech_rep}, several socio-demographic attributes exhibit meaningful geographical patterns. Race displays a noticeable degree of separation, with “While’’ and “Non-white’’ annotators forming distinct regions in the embedding space. In the plot for education level, the “Associate'' subgroup forms a more compact and distinguishable cluster compared to other categories. Income groups exhibit partial separation as well, suggesting that socio-economic background is correlated with differences in annotators’ perspectives on hate speech in detectable ways.

\paragraph{Toxic Dataset} In \cref{fig:toxic_rep}, annotators who consider religion to be important form a more compact region in the vector space. Education level also shows a clear pattern, with annotators holding Associate, Some College, Bachelor’s, and Master’s degrees arranged in a roughly left-to-right progression. Parental status is also associated with annotators’ perspectives on toxic content as shown in \cref{fig:toxic_rep_appendix} in \cref{sec:appendix_visualization}.



\subsection{Statistical Analysis of Representation}\label{sec:statistical_analysis_of_vectors}

While the visualizations provide an intuitive view of how socio-demographic attributes relate to the learned representations, dimensionality reduction inevitably discards some information. Moreover, for features with many subcategories (e.g., age groups, \cref{fig:hatesppeech_rep_appendix} and \cref{fig:toxic_rep_appendix} in \cref{sec:appendix_visualization}), the resulting plots become difficult to interpret. To complement the visualization results, we quantitatively analyze the structure of socio-demographic representations.

For each annotator, we measure the probability that its nearest neighbors share the same socio-demographic attribute as the selected annotator vector, computing the ratio of (i) the probability observed in the learned representation space, and (ii) the probability expected by chance.  
We apply bootstrap sampling with replacement (1,000 iterations) over annotators and obtain the statistics.

\paragraph{Observed Probability}
For each annotator vector \(i\), we retrieve its \(k = 50\) nearest neighbors and compute the proportion that share the same socio-demographic attribute. We then average this value across all annotators, expressed as:

\begin{equation}
\footnotesize
P_{\text{obs}} =
\frac{1}{N} \sum_{i=1}^{N}
\frac{
    \bigl|\{j \in \text{neighbors of } i : \text{attr}_j = \text{attr}_i \}\bigr|
}{k}
\end{equation}

\noindent where $N$ is the total number of annotators.
\vspace{4pt}
\paragraph{Expected Probability by Chance} \cite{mcpherson2001birds}
If annotator positions are random in the space, the probability that sampled annotators belong to the same category \(c\) is determined solely by its relative frequency. The expected probability that two randomly selected annotators fall into the same category by chance is:

\begin{equation}
\footnotesize
P_{\text{chance}} = \sum_{c \in C} \Pr(\text{annotator in category } c)^2
\end{equation}

\noindent where \(C\) is the set of socio-demographic categories, and \(c\) indexes each subcategory of \(C\). 

\paragraph{Homophily Ratio}
To quantify the degree of socio-demographic pattern in the learned space, we compute:
\begin{equation}
\footnotesize
\text{Ratio} = \frac{P_{\text{obs}}}{P_{\text{chance}}}.
\end{equation}

\noindent A Ratio > 1 indicates that the same socio-demographic attributes are more clustered than expected by chance, while a Ratio $\approx 1$ indicates random mixing of subcategories of a socio-demographic attribute.

As shown in \cref{tab:homophily_analysis}, education exhibits the highest ratio in both datasets, followed by ideology, income, and age group, all with ratios greater than 2. This indicates that these socio-demographic attributes are strongly associated with variation in annotator perspectives. Most socio-demographic features have ratios above 1, suggesting that they contribute to modeling hate speech and toxicity judgments, albeit to varying degrees. Among all features, the observed probability for “transgender” is lower than expected by chance. This may be due to the limited number of transgender annotators (fewer than 5\% of the annotator pool), resulting in patterns that are not well captured.


\section{Discussion: Socio-Demographic Specific Modeling for Bias Mitigation}

Prior studies \cite{cabitza2023toward,zhang2025proposal,feng-etal-2024-modular} argue that standard models trained on a single aggregated label ignore minority perspectives and encourage an averaged opinion derived from the training data. In this section, we discuss whether the proposed socio-contrastive method mitigates minority-ignorance bias and improves the fairness of representing opinions from underrepresented groups.
\cref{group_results} presents results across socio-demographic groups.

At first glance, there is no significant performance difference between the majority and minority groups under our method. For example, the F\textsubscript{1} scores for the female (0.735), male (0.711), and non-binary (0.711) groups on the hate speech task are comparable. On the toxicity task, the non-binary group achieves an F\textsubscript{1} score of 0.685, exceeding those of the female (0.640) and male (0.651) groups. Similarly, the transgender group achieves a score of 0.904, outperforming the non-transgender group (0.637).

However, we caution against overinterpreting these results and do not attribute this phenomenon to the proposed socio-demographic–enriched method, although the model has the potential to improve fairness and mitigate bias arising from the underrepresentation of minority opinions.
The test sizes for minority groups are very small, as minority groups constitute a smaller proportion of the annotator pool. In both datasets, each annotator labels only around 20 items on average. The number of test instances annotated by non-binary annotators is 64 for the hate speech task and 86 for the toxicity task, compared to over 10,000 instances for the female and male groups combined. As a result, these findings may be statistically unreliable. For example, test items annotated by non-binary annotators may happen to be easy (clearly hateful or non-hateful) or difficult to predict.

On the other hand, a specific socio-demographic group can be particularly sensitive to items that target or attack their own group. This may not be sufficiently represented in the current setting due to random item assignment in crowdsourced annotation. To more reliably assess whether minority group opinions are represented to a similar extent as majority groups, a better annotation procedure is required. All annotators should be asked to label a shared set of items containing potentially disputed cases.

\section{Conclusion}


This study indicates that incorporating socio-demographic attributes improves model performance in hate speech and toxic content classification. Our proposed Socio-Contrastive Learning outperforms models that do not use socio-demographic features and models that incorporate them via concatenation of socio-demographic and textual representations. Furthermore, we analyze the relationship between the learned socio-demographic representations and annotator perspectives. The results show that several socio-demographic factors are associated with variations in hate speech and toxicity-related perspectives to different extents.

\section*{Limitations}

Our investigation into the role of socio-demographic features in perspective modeling is subject to several constraints. First, large-scale datasets that are annotated by diverse populations and include rich, reliable socio-demographic metadata remain scarce. This prevents us from experimenting on broader and more diverse datasets.
Second, our analysis is restricted by the set of socio-demographic attributes provided by the original datasets. Although these attributes provide an initial lens on variation across groups, additional socio-demographic dimensions remain unexplored if they are potentially relevant to perspective differences.
Finally, for many data collection efforts, particularly those involving more “objective” tasks, socio-demographic information is typically not collected under the assumption that such tasks are unaffected by demographic factors. As a result, we are unable to systematically compare how the contribution of socio-demographic features to perspective modeling varies across task types. 

\bibliography{custom}

\newpage
\appendix

\section{Contrastive Loss in a Batch}\label{contrastive_loss_algorithm}

\begin{tcolorbox}[myalgo]
\begin{algorithm}[H]
\caption{\small \vspace{0.3em}\\{Contrastive Loss for Socio-Demographic Representation}\vspace{0.3em}}

\begin{algorithmic}[1]
\small 
\vspace{1.5em}
\Statex \textbf{Input:} Socio-Demo Embeddings 
 $\mathbf{E}\in\mathbb{R}^{B\times d}$, \vspace{0.3em}
\Statex \hspace{27pt}Annotator Labels $\mathbf{y}\in\mathbb{R}^B$, \vspace{0.3em}
\Statex \hspace{27pt}Text Identifiers $\mathbf{t}\in\mathbb{R}^B$, \vspace{0.3em}
\Statex \hspace{27pt}Temperature $\tau$.
\vspace{1.5em}

\Statex \textbf{Output:} Loss value $\mathcal{L}$

\vspace{1.5em}
\State $B \gets$ batch size

\vspace{1.5em}
\State Compute similarity matrix of socio-demo embeddings: 
\vspace{0.5em}

\State $\displaystyle \mathbf{S} \gets \frac{\mathbf{E}\mathbf{E}^\top}{\tau}$

\vspace{1.5em}
\State Compute text-matching mask: 
\State $\displaystyle \mathbf{M}_{\text{text}}[i,j] =
\begin{cases} 1 & t_i=t_j \\ 0 & \text{otherwise} \end{cases}$

\vspace{1.5em}
\State Positive mask (same text, same label): 
\vspace{0.5em}

\State $\displaystyle \mathbf{M}_{\text{pos}} \gets \mathbf{M}_{\text{text}} \cdot \mathbb{I}(y_i = y_j)$

\vspace{1.5em}
\State Remove diagonal: 
\vspace{0.5em}

\State $\displaystyle \mathbf{M}_{\text{pos}} \gets \mathbf{M}_{\text{pos}} \cdot (1 - \mathbf{I})$

\vspace{1.5em}
\State Loss for positive cases: 
\vspace{0.5em}

\State $\displaystyle \mathcal{L}_{\text{pos}} \gets - \frac{\sum \log \mathrm{Softmax}(\mathbf{S}) \cdot \mathbf{M}_{\text{pos}}}{\max(\sum \mathbf{M}_{\text{pos}},1)}$

\vspace{1.5em}
\State Negative mask (same text, different label): 
\vspace{0.5em}
\State $\displaystyle \mathbf{M}_{\text{neg}} \gets \mathbf{M}_{\text{text}} \cdot \mathbb{I}(y_i \neq y_j)$

\vspace{1.5em}

\State Loss for negative cases: 
\vspace{0.5em}
\State $\displaystyle \mathcal{L}_{\text{neg}} \gets \frac{\sum \mathrm{Softmax}(\mathbf{S}) \cdot \mathbf{M}_{\text{neg}}}{\max(\sum \mathbf{M}_{\text{neg}},1)}$

\vspace{1.5em}
\State \textbf{return} $\mathcal{L} = \mathcal{L}_{\text{pos}} + \mathcal{L}_{\text{neg}}$

\end{algorithmic}
\end{algorithm}
\end{tcolorbox}

\newpage

\section{Model Parameters}\label{parameter}

\begin{table}[h]
\renewcommand{\arraystretch}{1.3}
\centering
\small
\begin{tabular}{lr}
\toprule
\textbf{Hyperparameter} & \textbf{Value} \\
\midrule
\multicolumn{2}{l}{\textbf{Model Configuration}} \\
Hidden dimension (layer 1) & 512 \\
Hidden dimension (layer 2) & 256 \\
Socio-Contrastive Layers & (64, 128) \\
Dropout rate & 0.2 \\
Activation function &  ReLU \\
\hdashline
\multicolumn{2}{l}{\textbf{Training Configuration}} \\
Learning rate & 0.01 \\
Batch size & 32 \\
Number of epochs & 7 \\
Optimizer & Adam \\
\hdashline
\multicolumn{2}{l}{\textbf{Evaluation Settings}} \\
Decision threshold & 0.5 \\
F1 averaging & Binary \\
\bottomrule
\end{tabular}
\caption{Hyperparameter settings.}
\label{hyperparameter}
\end{table}

\onecolumn

\section{Visualization of Additional Socio-demographic Attributes}\label{sec:appendix_visualization}

\begin{figure*}[!h]
    \centering
    \includegraphics[width=0.9\linewidth]{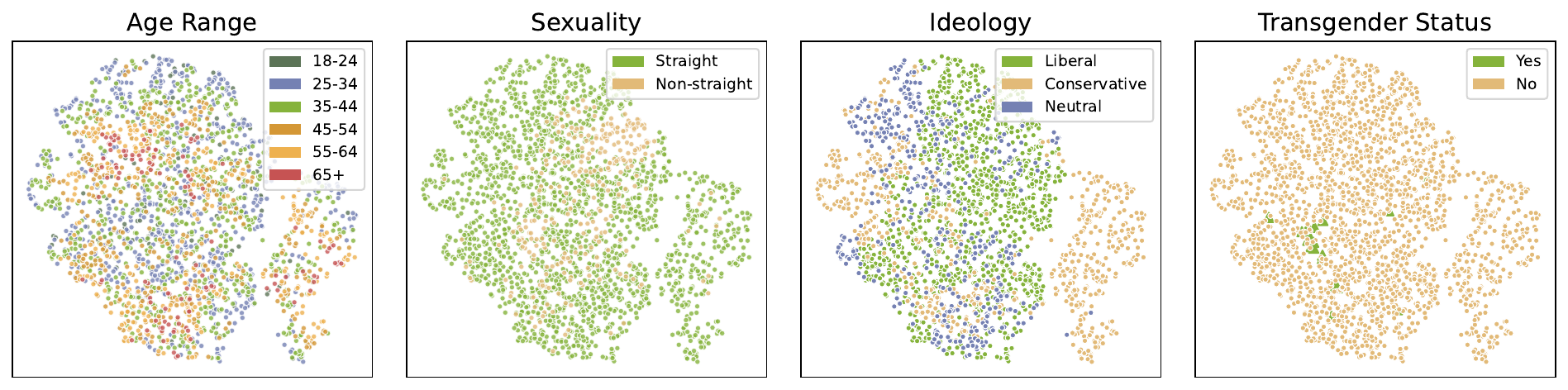}
    \caption{Visualization of Socio-Demographic Representation for the Hate Speech Dataset}
        \label{fig:hatesppeech_rep_appendix}
\end{figure*}

\begin{figure*}[!h]
    \centering
    \includegraphics[width=0.9\linewidth]{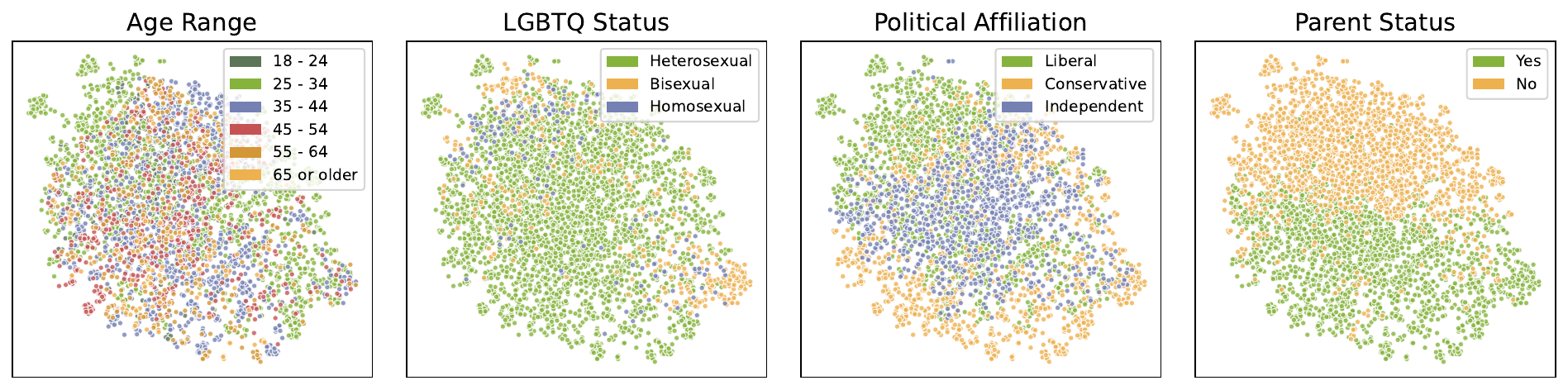}
    \caption{Visualization of Socio-Demographic Representation for the Toxic Dataset}
    \label{fig:toxic_rep_appendix}
\end{figure*}

\section{Additional Results}\label{bert_robert_results}

\begin{table*}[!h]
\centering
\renewcommand{\arraystretch}{1.2}
\resizebox{\textwidth}{!}{%
\begin{tabular}{lcccccc}
\toprule
\multirow{2}{*}{\textbf{BERT Encoder}}& \multicolumn{3}{c}{\textbf{Hate Speech}} 
& \multicolumn{3}{c}{\textbf{Toxic}} \\ \cmidrule(lr){2-4} \cmidrule(lr){5-7}

& {Precision} & {Recall} & {F1}
& {Precision} & {Recall} & {F1} \\ \hline
Simple Model
& 0.668 {\scriptsize$\pm$ 0.028}
& 0.735 {\scriptsize$\pm$ 0.065}
& 0.698 {\scriptsize$\pm$ 0.031}
& 0.565 {\scriptsize$\pm$ 0.015}
& 0.628 {\scriptsize$\pm$ 0.061}
& 0.593 {\scriptsize$\pm$ 0.019} \\

Socio Multi-Hot
& 0.736 {\scriptsize$\pm$ 0.011}
& 0.654 {\scriptsize$\pm$ 0.030}
& 0.692 {\scriptsize$\pm$ 0.019}
& 0.646 {\scriptsize$\pm$ 0.011}
& 0.562 {\scriptsize$\pm$ 0.031}
& 0.601 {\scriptsize$\pm$ 0.014} \\

Socio Embedding
& 0.740 {\scriptsize$\pm$ 0.013}
& 0.629 {\scriptsize$\pm$ 0.033}
& 0.679 {\scriptsize$\pm$ 0.016}
& 0.655 {\scriptsize$\pm$ 0.013}
& 0.546 {\scriptsize$\pm$ 0.036}
& 0.595 {\scriptsize$\pm$ 0.016} \\

Socio Contrastive (Ours)
& 0.690 {\scriptsize$\pm$ 0.038}
& 0.736 {\scriptsize$\pm$ 0.092}
& 0.708 {\scriptsize$\pm$ 0.026}
& 0.652 {\scriptsize$\pm$ 0.012}
& 0.608 {\scriptsize$\pm$ 0.039}
& 0.629 {\scriptsize$\pm$ 0.016} \\

\bottomrule
\end{tabular}}

\caption{Performance comparison of models (mean ± std). Text representations are encoded using BERT.}
\label{appendix_result_table_bert}

\end{table*}

\begin{table*}[!h]
\centering
\renewcommand{\arraystretch}{1.2}
\resizebox{\textwidth}{!}{%
\begin{tabular}{lcccccc}
\toprule
\multirow{2}{*}{\textbf{RoBERTa Encoder}}& \multicolumn{3}{c}{\textbf{Hate Speech}} 
& \multicolumn{3}{c}{\textbf{Toxic}} \\ \cmidrule(lr){2-4} \cmidrule(lr){5-7}

& {Precision} & {Recall} & {F1}
& {Precision} & {Recall} & {F1} \\ \hline
Simple Model
& 0.683 {\scriptsize$\pm$ 0.017}
& 0.724 {\scriptsize$\pm$ 0.049}
& 0.702 {\scriptsize$\pm$ 0.017}
& 0.569 {\scriptsize$\pm$ 0.017}
& 0.670 {\scriptsize$\pm$ 0.063}
& 0.613 {\scriptsize$\pm$ 0.019} \\

Socio Multi-Hot
& 0.742 {\scriptsize$\pm$ 0.009}
& 0.610 {\scriptsize$\pm$ 0.059}
& 0.668 {\scriptsize$\pm$ 0.033}
& 0.650 {\scriptsize$\pm$ 0.018}
& 0.598 {\scriptsize$\pm$ 0.055}
& 0.621 {\scriptsize$\pm$ 0.023} \\

Socio Embedding
& 0.749 {\scriptsize$\pm$ 0.008}
& 0.650 {\scriptsize$\pm$ 0.031}
& 0.696 {\scriptsize$\pm$ 0.017}
& 0.647 {\scriptsize$\pm$ 0.009}
& 0.606 {\scriptsize$\pm$ 0.026}
& 0.625 {\scriptsize$\pm$ 0.009} \\

Socio Contrastive (Ours)
& 0.677 {\scriptsize$\pm$ 0.074}
& 0.768 {\scriptsize$\pm$ 0.112}
& 0.710 {\scriptsize$\pm$ 0.029}
& 0.645 {\scriptsize$\pm$ 0.013}
& 0.646 {\scriptsize$\pm$ 0.029}
& 0.645 {\scriptsize$\pm$ 0.010} \\

\bottomrule
\end{tabular}}
\caption{Performance comparison of models (mean ± std). Text representations are encoded using RoBERTa.}
\label{appendix_result_table_RoBERTa}

\end{table*}

\vspace{82pt}

\section{Results by Socio-Demographic Group Divisions}\label{group_results}

\begin{table}[!h]
\renewcommand{\arraystretch}{1.2}
\resizebox{\textwidth}{!}{%
\begin{tabular}{@{}lccccc@{}}
\toprule
             \textbf{Group Division (n = Test Size)}               & \textbf{Simple Model}      & \textbf{Multi-Task}      & \textbf{Socio Multi-Hot}     & \textbf{Socio Embedding}      & \textbf{Socio Contrastive} \\ \midrule
{\cellcolor[HTML]{849F7F}\textbf{Gender}}                                                                                                                                             \\ 
female (n=6018)                                     & 0.427                      & 0.614                    & 0.698                        & 0.706                         & 0.735                      \\
male (n=4225)                                       & 0.398                      & 0.601                    & 0.675                        & 0.688                         & 0.711                      \\
non-binary (n=64)                                   & 0.418                      & 0.637                    & 0.596                        & 0.744                         & 0.711                      \\ \hdashline
{\cellcolor[HTML]{849F7F}\textbf{{Age Group}}}                                                                                                                                         \\ 
26-35 (n=3565)                                      & 0.405                      & 0.593                    & 0.672                        & 0.669                         & 0.707                      \\
36-45 (n=2642)                                      & 0.417                      & 0.604                    & 0.672                        & 0.686                         & 0.722                      \\
46-55 (n=1702)                                      & 0.415                      & 0.634                    & 0.721                        & 0.738                         & 0.758                      \\
56-65 (n=1013)                                      & 0.426                      & 0.637                    & 0.722                        & 0.754                         & 0.758                      \\
0-25 (n=1001)                                       & 0.424                      & 0.574                    & 0.622                        & 0.646                         & 0.662                      \\
66+ (n=419)                                         & 0.435                      & 0.629                    & 0.767                        & 0.756                         & 0.778                      \\ \hdashline
{\cellcolor[HTML]{849F7F}\textbf{Education Level}}                                                                                                                                    \\ 
{college\_grad\_ba (n=3909)}      & 0.415                      & 0.604                    & 0.666                        & 0.675                         & 0.716                      \\
some\_college (n=2517)                              & 0.415                      & 0.594                    & 0.669                        & 0.687                         & 0.720                      \\
{college\_grad\_aa (n=1506)}      & 0.433                      & 0.624                    & 0.732                        & 0.733                         & 0.741                      \\
masters (n=999)                                     & 0.395                      & 0.611                    & 0.680                        & 0.695                         & 0.719                      \\
{high\_school\_grad (n=946)}      & 0.393                      & 0.617                    & 0.729                        & 0.751                         & 0.761                      \\
{professional\_degree (n=244)}    & 0.425                      & 0.660                    & 0.745                        & 0.752                         & 0.757                      \\
phd (n=149)                                         & 0.381                      & 0.602                    & 0.655                        & 0.715                         & 0.652                      \\
{some\_high\_school (n=89)}       & 0.500                      & 0.641                    & 0.731                        & 0.637                         & 0.676                      \\\hdashline
{\cellcolor[HTML]{849F7F}\textbf{Sexuality Straight}}                                                                                                                                 \\

Yes (n=9020)                                        & 0.413                      & 0.608                    & 0.690                        & 0.700                         & 0.727                      \\
No (n=1339)                                         & 0.424                      & 0.608                    & 0.667                        & 0.693                         & 0.715                     
  \\\hdashline
{\cellcolor[HTML]{849F7F}\textbf{Race White}}                                                                                                                                         \\
Yes (n=8295)                                        & 0.418                      & 0.609                    & 0.696                        & 0.707                         & 0.731                      \\
No (n=2064)                                         & 0.402                      & 0.605                    & 0.653                        & 0.667                         & 0.699                      \\\hdashline
{\cellcolor[HTML]{849F7F}\textbf{Political Ideology}}                                                                                                                                 \\
liberal (n=2507)                                    & 0.406                      & 0.609                    & 0.690                        & 0.698                         & 0.730                      \\
neutral (n=1838)                                    & 0.409                      & 0.608                    & 0.679                        & 0.699                         & 0.720                      \\
slightly\_liberal (n=1637)                          & 0.425                      & 0.607                    & 0.669                        & 0.684                         & 0.704                      \\
conservative (n=1321)                               & 0.413                      & 0.618                    & 0.699                        & 0.704                         & 0.747                      \\
{slightly\_conservative (n=1321)} & 0.439                      & 0.590                    & 0.677                        & 0.703                         & 0.712                      \\
{extremely\_liberal (n=1185)}     & 0.393                      & 0.601                    & 0.678                        & 0.689                         & 0.721                      \\
{extremely\_conservative (n=315)} & 0.427                      & 0.632                    & 0.747                        & 0.728                         & 0.765                      \\\hdashline
{\cellcolor[HTML]{849F7F}\textbf{Income Range}}                                                                                                                                       \\
10k-50k (n=4264)                                    & 0.415                      & 0.607                    & 0.679                        & 0.689                         & 0.718                      \\
50k-100k (n=4034)                                   & 0.425                      & 0.614                    & 0.695                        & 0.719                         & 0.732                      \\
100k-200k (n=1299)                                  & 0.410                      & 0.595                    & 0.693                        & 0.664                         & 0.730                      \\
\textless{}10k (n=555)                              & 0.371                      & 0.611                    & 0.719                        & 0.718                         & 0.751                      \\
\textgreater{}200k (n=200)                          & 0.318                      & 0.577                    & 0.513                        & 0.564                         & 0.598                      \\ \bottomrule
\end{tabular}}
\caption{F1 Scores for Socio-Demographic Group Divisions on the Hate Speech Task.}\label{tab:group_specific_hate}
\end{table}

\begin{table}[]
\renewcommand{\arraystretch}{1.2}
\resizebox{\textwidth}{!}{%
\begin{tabular}{@{}lccccc@{}}
\toprule
                                \textbf{Group Division (n = Test Size)}                                              & \textbf{Simple Model} & \textbf{Multi-Task} & \textbf{Socio Multi-Hot} & \textbf{Socio Embedding} & \textbf{Socio Contrastive} \\ \midrule
\textbf{Gender}                                                                 &                       &                     &                          &                          &                            \\
Female (n=8515)                                                                 & 0.485                 & 0.527               & 0.618                    & 0.623                    & 0.640                      \\
Male (n=7428)                                                                   & 0.485                 & 0.562               & 0.615                    & 0.634                    & 0.651                      \\
Prefer not to say (n=114)                                                       & 0.523                 & 0.511               & 0.535                    & 0.534                    & 0.624                      \\
Nonbinary (n=86)                                                                & 0.564                 & 0.500               & 0.663                    & 0.706                    & 0.685                      \\\hdashline
\textbf{Age Group}                                                              &                       &                     &                          &                          &                            \\
25 - 34 (n=6246)                                                                & 0.500                 & 0.582               & 0.641                    & 0.647                    & 0.668                      \\
35 - 44 (n=4205)                                                                & 0.471                 & 0.528               & 0.585                    & 0.604                    & 0.609                      \\
45 - 54 (n=2453)                                                                & 0.487                 & 0.526               & 0.605                    & 0.611                    & 0.646                      \\
18 - 24 (n=1425)                                                                & 0.470                 & 0.503               & 0.635                    & 0.640                    & 0.664                      \\
55 - 64 (n=1155)                                                                & 0.485                 & 0.480               & 0.601                    & 0.623                    & 0.627                      \\
65 or older (n=609)                                                             & 0.464                 & 0.495               & 0.575                    & 0.589                    & 0.588                      \\
Prefer not to say (n=58)                                                        & 0.558                 & 0.645               & 0.628                    & 0.746                    & 0.653                      \\\hdashline
\textbf{Education}                                                              &                       &                     &                          &                          &                            \\
Bachelor's degree in college (n=6199)                                  & 0.488                 & 0.576               & 0.621                    & 0.630                    & 0.656                      \\
Some college but no degree (n=3282)                                             & 0.458                 & 0.461               & 0.582                    & 0.584                    & 0.600                      \\
Master's degree (n=2436)                                                        & 0.538                 & 0.655               & 0.719                    & 0.736                    & 0.738                      \\
Associate degree in college (2-year) (n=1997)                                   & 0.466                 & 0.427               & 0.546                    & 0.548                    & 0.596                      \\
High school graduate (n=1500) & 0.455                 & 0.486               & 0.559                    & 0.578                    & 0.560                      \\
Professional degree (JD, MD) (n=251)                                            & 0.526                 & 0.440               & 0.542                    & 0.571                    & 0.603                      \\
Doctoral degree (n=250)                                                         & 0.466                 & 0.426               & 0.451                    & 0.537                    & 0.525                      \\
Less than high school degree (n=138)                                            & 0.488                 & 0.576               & 0.661                    & 0.643                    & 0.653                      \\
Prefer not to say (n=60)                                                        & 0.509                 & 0.647               & 0.641                    & 0.619                    & 0.665                      \\
Other (n=38)                                                                    & 0.544                 & 0.553               & 0.302                    & 0.665                    & 0.625                      \\\hdashline
\textbf{Sexuality}                                                              &                       &                     &                          &                          &                            \\
Heterosexual (n=13730)                                                          & 0.478                 & 0.522               & 0.594                    & 0.605                    & 0.626                      \\
Bisexual (n=1305)                                                               & 0.551                 & 0.713               & 0.775                    & 0.790                    & 0.789                      \\
Homosexual (n=611)                                                              & 0.524                 & 0.580               & 0.670                    & 0.696                    & 0.687                      \\
Prefer not to say (n=331)                                                       & 0.471                 & 0.536               & 0.633                    & 0.652                    & 0.645                      \\
Other (n=126)                                                                   & 0.451                 & 0.343               & 0.493                    & 0.486                    & 0.540                      \\\hdashline
\textbf{Transgender}                                                            &                       &                     &                          &                          &                            \\
No (n=15709)                                                                    & 0.482                 & 0.534               & 0.607                    & 0.619                    & 0.637                      \\
Yes (n=252)                                                                     & 0.635                 & 0.851               & 0.893                    & 0.921                    & 0.904                      \\
Prefer not to say (n=190)                                                       & 0.543                 & 0.534               & 0.675                    & 0.654                    & 0.701                      \\\hdashline
\textbf{Political Affiliation}                                                  &                       &                     &                          &                          &                            \\
Liberal (n=6551)                                                                & 0.482                 & 0.490               & 0.577                    & 0.587                    & 0.620                      \\
Conservative (n=4383)                                                           & 0.497                 & 0.612               & 0.660                    & 0.677                    & 0.679                      \\
Independent (n=4197)                                                            & 0.479                 & 0.533               & 0.612                    & 0.613                    & 0.641                      \\
Prefer not to say (n=661)                                                       & 0.507                 & 0.552               & 0.694                    & 0.716                    & 0.688                      \\
Other (n=359)                                                                   & 0.454                 & 0.556               & 0.537                    & 0.598                    & 0.569                      \\\hdashline
\textbf{Religion Importance}                                                    &                       &                     &                          &                          &                            \\
Not important (n=5848)                                                          & 0.428                 & 0.424               & 0.511                    & 0.505                    & 0.554                      \\
Very important (n=4803)                                                         & 0.525                 & 0.629               & 0.676                    & 0.697                    & 0.702                      \\
Somewhat important (n=3354)                                                     & 0.511                 & 0.587               & 0.654                    & 0.680                    & 0.685                      \\
Not too important (n=1865)                                                      & 0.490                 & 0.463               & 0.596                    & 0.566                    & 0.617                      \\
Prefer not to say (n=281)                                                       & 0.486                 & 0.550               & 0.661                    & 0.660                    & 0.635                      \\\hdashline
\textbf{Is Parent}                                                              &                       &                     &                          &                          &                            \\
Yes (n=8268)                                                                    & 0.510                 & 0.581               & 0.652                    & 0.668                    & 0.672                      \\
No (n=7701)                                                                     & 0.459                 & 0.494               & 0.571                    & 0.578                    & 0.615                      \\
Prefer not to say (n=182)                                                       & 0.402                 & 0.497               & 0.531                    & 0.456                    & 0.561                   \\  \bottomrule
\end{tabular}}
\caption{F1 Scores for Socio-Demographic Group Divisions on the Toxic Content Task.}\label{tab:group_division_toxic}
\end{table}

\twocolumn

\end{document}